\documentclass[runningheads]{llncs}
\usepackage[T1]{fontenc}
\usepackage{graphicx,verbatim}
\usepackage{amsmath}
\usepackage{tabularx}
\usepackage{multirow}
\usepackage{subcaption}
\begin{document}
\title{A methodology for clinically driven interactive segmentation evaluation}
%

\author{Parhom Esmaeili\inst{1} \and Pedro Borges\inst{1} \and Virginia Fernandez\inst{1} \and Eli Gibson\inst{2} \and Sebastien Ourselin\inst{1} \and M. Jorge Cardoso\inst{1}}  

\authorrunning{P. Esmaeili et al.}
\institute{School of Biomedical Engineering and Imaging Sciences, KCL, UK
\and Siemens Healthineers, Princeton NJ, USA \\
    \email{parhom.esmaeili@kcl.ac.uk}}

\maketitle          
\begin{abstract}

Interactive segmentation is a promising strategy for building robust, generalisable algorithms for volumetric medical image segmentation. However, inconsistent and clinically unrealistic evaluation hinders fair comparison and misrepresents real-world performance. We propose a clinically grounded methodology for defining evaluation tasks and metrics, and built a software framework for constructing standardised evaluation pipelines. We evaluate state-of-the-art algorithms across heterogeneous and complex tasks and observe that \textbf{(i)} minimising information loss when processing user interactions is critical for model robustness, \textbf{(ii)} adaptive-zooming mechanisms boost robustness and speed convergence, \textbf{(iii)} performance drops if validation prompting behaviour/budgets differ from training, \textbf{(iv)} 2D methods perform well with slab-like images and coarse targets, but 3D context helps with large or irregularly shaped targets, \textbf{(v)} performance of non-medical-domain models (e.g. SAM2) degrades with poor contrast and complex shapes.


\end{abstract}
\section{Introduction}

Segmentation of anatomical structures and pathologies is essential for extracting imaging biomarkers and delineating structures for use in treatment planning, patient monitoring, and guided therapies. However, manual segmentation of imaging volumes remains time-consuming and bottlenecks clinical workflows \cite{Lenchik2019AutomatedReview}. While automatic segmentation algorithms trained via supervised learning have demonstrated strong performance on many targets \cite{Isensee2021NnU-Net:Segmentation}, fine structures, heterogeneous and out-of-distribution targets/modalities can be particularly hard. Moreover, the limited availability of annotated data, caused by data sharing restrictions \cite{deKok2023ARegulation} and laboriousness of manual annotation often hinders automated segmentation performance. Algorithms with strong zero-shot generalisation and semi-automated refinement, especially if they can improve on tasks with repeated exposure and reach expert-level performance, would provide substantial clinical impact. Interactive segmentation incorporates user prompts to guide and refine segmentations, helping to address many of these challenges by reducing reliance on image-derived features. Classical interactive methods relying on low-level image features \cite{Kass1988Snakes:Models,Boykov2001InteractiveImages} are already integrated into segmentation tools such as ITK-Snap and 3D Slicer \cite{Yushkevich2006User-guidedReliability,Fedorov20123DNetwork} but often struggle with low-contrast or complex tissue boundaries. Recently, deep learning–based interactive segmentation has advanced in natural \cite{Kirillov2023SegmentAnything,Ravi2024SAMVideos} and medical imaging domains \cite{Cheng2023SAM-Med2D,Wang2023SAM-Med3D,Du2024SegVol:Segmentation,Isensee2025NnInteractive:Segmentation,Li2024Prism:Prompts}. However, current approaches to validate and compare interactive segmentation algorithms are flawed, preventing fair assessment and overestimating progress. This paper outlines key pitfalls, offers guidelines, proposes a framework for constructing evaluation pipelines, and analyses prominent algorithms using these.

\section{Validation Pitfalls}\label{Section.2 Pitfalls}

In a clinical setting, interactive segmentation algorithms are AI-assistants that translate user inputs and prompts into model-ready formats and return segmentations in the original image space. However, most algorithmic validation experiments fail to appropriately standardise the representation of the inputs, prompts and outputs by introducing pre-processing steps such as: \textbf{(i)} resampling to model-specific resolutions for prompting and validation \cite{Du2024SegVol:Segmentation,Wang2023SAM-Med3D}; or \textbf{(ii)} simulating prompts and calculating metrics on restrictive image subregions (e.g., slices or annotation-informed image crops) \cite{Cheng2023SAM-Med2D,Wang2023SAM-Med3D,Li2024Prism:Prompts}. Since the number of interactions and inference times are often used as a efficiency proxy, reporting metrics on a sub-region \cite{Li2024Prism:Prompts} misrepresents annotation-effort and hampers comparison. Performing evaluation in the original image space (full resolution and field-of-view) is recommended to better reflect deployment realities.

\subsection{Task Taxonomies}\label{Section2.1:Taxonomies}

Here we outline a task taxonomy covering algorithmic axes of complexity; essential for clinicians to select interactive segmentation algorithms suited to their applications, and to aid researchers in identifying remaining challenges. We also suggest suitable evaluation principles, where applicable. 

\textbf{Problem 1:} Ulrich et al. \cite{Ulrich2025RadioActive:Benchmark} note that most studies focus on segmentation tasks easily handled by automated methods, neglecting clinically difficult cases. Challenging tasks include targets with ambiguous boundaries or heterogeneous appearance (e.g., tumours), very small targets (white matter lesions), geometrically complex or topologically constrained structures (like vascular networks), and multi-target segmentation where simple label merging is suboptimal (e.g., nested or adjacent semantic targets, or when spatial contiguity does not define separate instances). Interactive segmentation is also often evaluated as binary instance segmentation, even in multi-target cases, with benchmarks \cite{Ulrich2025RadioActive:Benchmark} relying on potentially unreliable assumptions of contiguity for identifying instances (e.g., touching lesions may be distinct metastases). Evaluations should prioritise challenging tasks, assess multi-target segmentation in parallel for tasks where simple labels merging is inadequate, and critically assess contiguity-based assumptions. 

\textbf{Problem 2:} Most studies omit multi-modal (e.g., PET-CT \cite{Gatidis2022ALesions}) or multi-sequence (e.g., mpMRI \cite{Maleki2025AnalysisMRI}) image data even when it provides essential clinical context. Validation pipelines should aim to include these, when applicable to the task, to foster development of more clinically impactful algorithms. 

\textbf{Problem 3:} Zero-shot models are mainly trained and tested in case independent and fully interactive settings. Evaluation tasks testing automatic initialisation techniques or model adaptation to repeated tasks (e.g. fine tuning, active learning) could foster algorithms which further expedite annotation. 

\textbf{Problem 4:} Prompting methods (points, scribbles, boxes, lassos) differ in effectiveness and annotation effort due to variation in implied intent and spatial constraints (e.g., lassos are laborious for jagged shapes). Also, some algorithms fully support all prompts during initialisation and editing \cite{Isensee2025NnInteractive:Segmentation}, while others only support a subset of prompts in specific circumstances \cite{Du2024SegVol:Segmentation,Cheng2023SAM-Med2D,Ravi2024SAMVideos}. For a fair evaluation one should consider using prompt configurations that are supported by all the algorithms being compared. User-effort should also be estimated using the number of interactions and also ideally the prompt placement effort as proxies, with placement efforts being estimated via user studies. This approach could also facilitate effort comparisons across prompting mechanisms. 
\subsection{Metrics}\label{Section:2.2 Metrics}
 

As recommended by Maier-Hein et al. \cite{Maier-Hein2024MetricsValidation}, evaluations should report complementary metrics that measure different aspects of the task, e.g. both overlap based metrics (e.g., Dice) and boundary-aware metrics (e.g., normalised surface Dice, NSD). Yet, many studies and even benchmarks \cite{Ulrich2025RadioActive:Benchmark} rely solely on Dice, which can misrepresent performance on small and fine, or geometrically complex structures. Performance variability during iterative refinement is also rarely captured, despite its importance; Dice and NSD area under the curve (AUC) metrics normalised by interaction count \cite{Antonov2024RClicks:Segmentation} can be used to measure convergence speed and stability. As AUC may penalise methods that require more interactions but that yield a better final outcome, especially under tight interaction budgets, assessing convergence to clinically meaningful endpoints is important. Existing benchmarks also only assume zero-shot settings for clinical use overlooking comparisons to strong automated methods \cite{Ulrich2025RadioActive:Benchmark}, such as nnU-Net \cite{Isensee2021NnU-Net:Segmentation}, which can offer lower annotator effort on repeated tasks. Evaluations should consider larger interaction budgets to identify emergent behaviours and use clinical criteria or task-specialist automatic baselines to determine termination criteria. The latter can also highlight any superiority of interactive methods in clinical workflows.

\section{Methods}\label{Section3:Methods}

\begin{figure}[t!]
\centering
\includegraphics[width=0.965\textwidth]{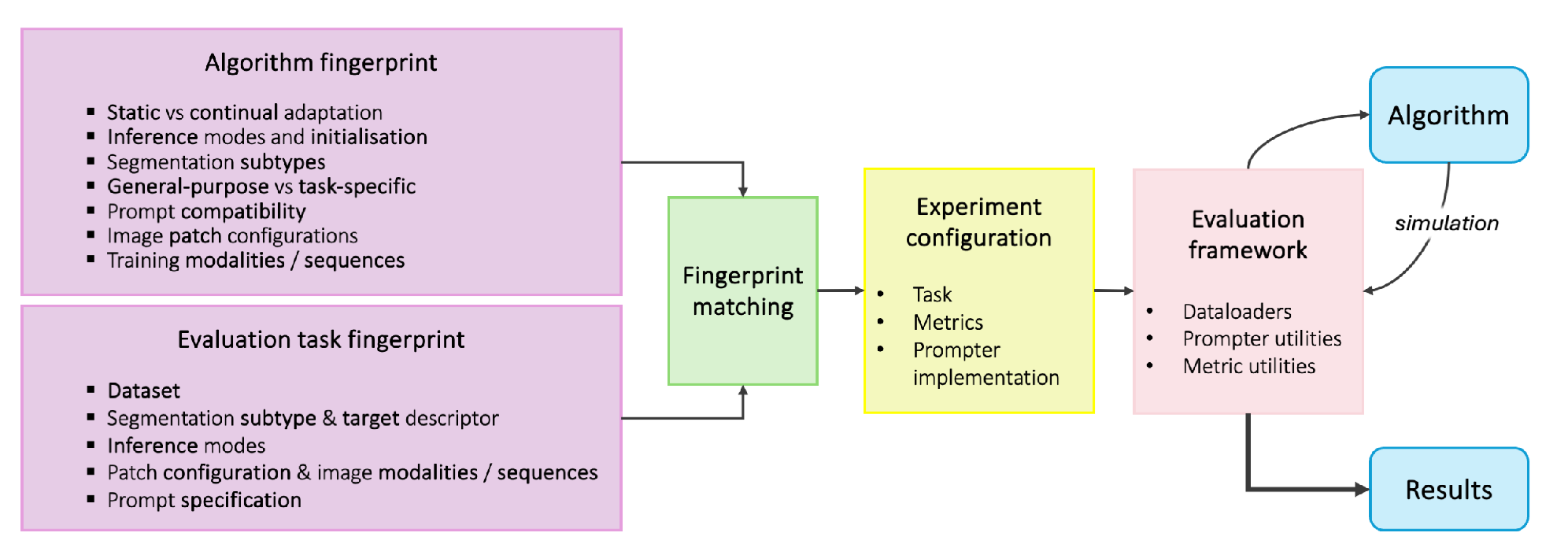}
\caption{A flowchart for identifying compatible experiments that are passed to the proposed evaluation framework for interaction simulation and evaluation.}
\label{fig:Eval_Framework}
\end{figure}

The interactive nature of these algorithms requires continuous communication between the algorithm and a validation framework. Therefore, we developed a modular framework that decouples the generation of segmentation requests (image patch, prompts, task description) and metric computation from the inference algorithm; and integrated this into a task selection pipeline (Fig.\ref{fig:Eval_Framework}). First, we characterise algorithms with characteristics, or fingerprints, spanning:\textbf{(i)} whether algorithms are static at inference or learn to adapt over repeated tasks; \textbf{(ii)} supported inference modes (initialisation strategies and whether editing is natively supported or partially compatible); \textbf{(iii)} segmentation subtypes supported natively (e.g., binary, multi-class, semantic, instance, or panoptic segmentation); \textbf{(iv)} general-purpose versus target-specific training; \textbf{(v)} compatibility across prompt types (points, scribbles, bounding boxes, lassos), and any constraints on their usage (see Section~\ref{Section2.1:Taxonomies}); \textbf{(vi)} model-native image patch configurations (voxel counts, channel count) and algorithmic adaptability to deviations from these; and \textbf{(vii)} the imaging modalities seen during training. These fingerprints need to be cross-referenced with candidate tasks, defined by: dataset, segmentation subtype and target, image patch configuration (voxel count, channel count, modality and sequence) and prompt specification to select compatible tasks. The task then determines the corresponding evaluation metrics, while the algorithm fingerprints define compatible prompting mechanisms.

\begin{figure}[t!] 
\centering
\includegraphics[width=0.95\textwidth]{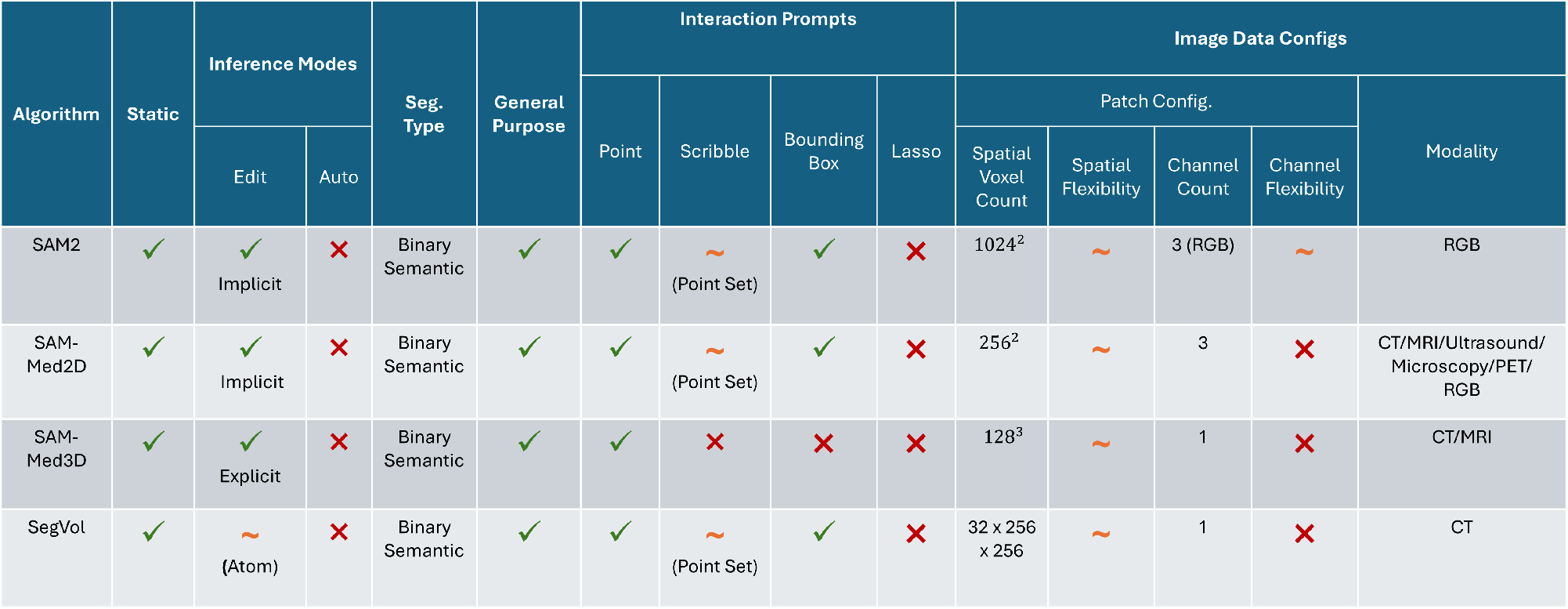}
\caption{Algorithm fingerprints. Ticks: full support; Tildes: partial support; Crosses: no support. Implicit editing uses full prompt memory, explicit only uses current prompts, and atomic editing re-does inference with full prompt memory. All except SAM-Med3D support scribbles (represented as a set of points); SAM-Med3D is limited by a one-point-per-class constraint. Only SAM2 natively supports multi-channel radiological images.
}
\label{fig:algo_fingerprint}
\end{figure}

We integrated SAM2 \cite{Ravi2024SAMVideos}, SAM-Med2D \cite{Cheng2023SAM-Med2D}, SAM-Med3D \cite{Wang2023SAM-Med3D}, and SegVol \cite{Du2024SegVol:Segmentation} into applications compatible with the proposed framework's segmentation request  definition (image-patch, corresponding prompt coordinates and the segmentation task description). Model fingerprints are shown in Fig.~\ref{fig:algo_fingerprint}, all applications were kept online throughout experimental sessions. For SAM2 and SAM-Med2D images intensities were clipped (0.5–99.5th percentiles) and min-max normalised to [0,255] on full volumes. Inference for 2D models is performed on axial slices with images and prompt coordinates remapped to native voxel counts. Outputs are then remapped to input voxel counts. No changes were needed for SAM-Med3D as it provided native normalisation and voxel count adaptation. SegVol lacked voxel count adaptation for points, so points were translated into relative coordinates for the cropped zoom-in region, and then remapped to native voxel counts for zoom-out inference. No algorithmic changes were made beyond those mentioned above. 

\section{Experiments and Results}

\begin{figure}[t!]
\caption{\textbf{Top:} Median Dice and NSD across the refinement simulations as image voxel count increases (hippocampus, brain tumour core, pancreas). \textbf{Bottom:} Summary metrics for experiment 1; all metrics but NoF report dataset medians, NoF reports percentages. Bold indicates best metric per task.}
\centering
\includegraphics[width=\textwidth]{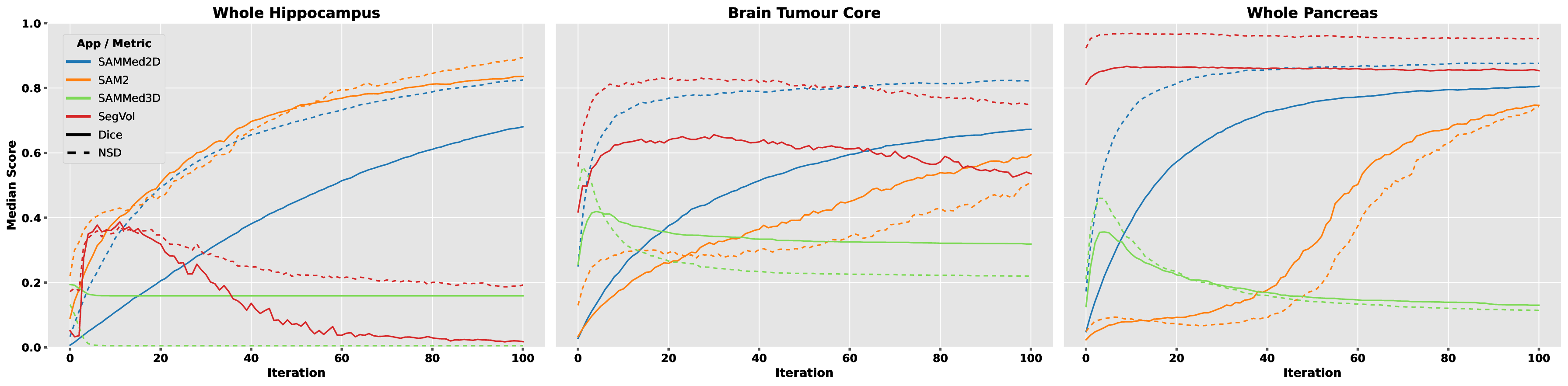}

\centering
\resizebox{0.8\textwidth}{!}{%
\begin{tabular}{|c|c|ccc|ccc|cc|}
\hline
\multirow{2}{*}{Task} &
  \multirow{2}{*}{Algorithm} &
  \multicolumn{3}{c|}{Dice} &
  \multicolumn{3}{c|}{NSD} &
  \multicolumn{2}{c|}{NoI} \\ \cline{3-10} 
 &
   &
  \multicolumn{1}{c|}{Init} &
  \multicolumn{1}{c|}{Iter. 100} &
  nAUC &
  \multicolumn{1}{c|}{Init} &
  \multicolumn{1}{c|}{Iter. 100} &
  nAUC &
  \multicolumn{1}{c|}{nNoI} &
  NoF \\ \hline
\multirow{4}{*}{Hippocampus} &
  SAM2 &
  \multicolumn{1}{c|}{0.090} &
  \multicolumn{1}{c|}{\textbf{0.836}} &
  \textbf{0.667}
   &
  \multicolumn{1}{c|}{\textbf{0.220}} &
  \multicolumn{1}{c|}{\textbf{0.894}} &
  \textbf{0.682}
   &
  \multicolumn{1}{c|}{1.0} &
  \textbf{95.4} \\ \cline{2-10} 
 &
  SAM-Med2D &
  \multicolumn{1}{c|}{0.007} &
  \multicolumn{1}{c|}{0.681} &
  0.415
   &
  \multicolumn{1}{c|}{0.033} &
  \multicolumn{1}{c|}{0.825} &
  0.630
    &
  \multicolumn{1}{c|}{1.0} &
   100 \\ \cline{2-10} 
 &
  SAM-Med3D &
  \multicolumn{1}{c|}{\textbf{0.194}} &
  \multicolumn{1}{c|}{0.159} &
   0.160
   &
  \multicolumn{1}{c|}{0.132} &
  \multicolumn{1}{c|}{0.005} &
    0.008
   &
  \multicolumn{1}{c|}{1.0} &
   100 \\ \cline{2-10} 
 &
  SegVol &
  \multicolumn{1}{c|}{0.051} &
  \multicolumn{1}{c|}{0.018} &
    0.159
   &
  \multicolumn{1}{c|}{0.172} &
  \multicolumn{1}{c|}{0.192} &
   0.257
   &
  \multicolumn{1}{c|}{1.0} &
   100 \\ \hline
\multirow{4}{*}{Brain Tumour} &
  SAM2 &
  \multicolumn{1}{c|}{0.034} &
  \multicolumn{1}{c|}{0.594} &
   0.396 &
  \multicolumn{1}{c|}{0.130} &
  \multicolumn{1}{c|}{0.509} &
   0.366
   &
  \multicolumn{1}{c|}{1.0} &
   55.8
   \\ \cline{2-10} 
 &
  SAM-Med2D &
  \multicolumn{1}{c|}{0.028} &
  \multicolumn{1}{c|}{\textbf{0.672}} &
   0.504
   &
  \multicolumn{1}{c|}{0.250} &
  \multicolumn{1}{c|}{\textbf{0.822}} &
   0.776
   &
  \multicolumn{1}{c|}{0.851} &
   43.8
   \\ \cline{2-10} 
 &
  SAM-Med3D &
  \multicolumn{1}{c|}{0.256} &
  \multicolumn{1}{c|}{0.319} &
   0.341
   &
  \multicolumn{1}{c|}{0.489} &
  \multicolumn{1}{c|}{0.219} &
  0.256
   &
  \multicolumn{1}{c|}{1.0} &
   77.9 \\ \cline{2-10} 
 &
  SegVol &
  \multicolumn{1}{c|}{\textbf{0.418}} &
  \multicolumn{1}{c|}{0.536} &
   \textbf{0.595}
   &
  \multicolumn{1}{c|}{\textbf{0.558}} &
  \multicolumn{1}{c|}{0.750} &
   \textbf{0.791}
   &
  \multicolumn{1}{c|}{\textbf{0.060}} &
   \textbf{33.7}
   \\ \hline
\multirow{4}{*}{Pancreas} &
  SAM2 &
  \multicolumn{1}{c|}{0.023} &
  \multicolumn{1}{c|}{0.746} &
   0.375
   &
  \multicolumn{1}{c|}{0.050} &
  \multicolumn{1}{c|}{0.743} &
   0.321
   &
  \multicolumn{1}{c|}{1.0} &
   75.1 \\ \cline{2-10} 
 &
  SAM-Med2D &
  \multicolumn{1}{c|}{0.049} &
  \multicolumn{1}{c|}{0.806} &
   0.674
   &
  \multicolumn{1}{c|}{0.173} &
  \multicolumn{1}{c|}{0.876} &
   0.824
   &
  \multicolumn{1}{c|}{1.0} &
   63.3 \\ \cline{2-10} 
 &
  SAM-Med3D &
  \multicolumn{1}{c|}{0.126} &
  \multicolumn{1}{c|}{0.130} &
   0.189
   &
  \multicolumn{1}{c|}{0.210} &
  \multicolumn{1}{c|}{0.114} &
   0.184
   &
  \multicolumn{1}{c|}{1.0} &
   98.9 \\ \cline{2-10} 
 &
  SegVol &
  \multicolumn{1}{c|}{\textbf{0.812}} &
  \multicolumn{1}{c|}{\textbf{0.854}} &
   \textbf{0.857}
   &
  \multicolumn{1}{c|}{\textbf{0.924}} &
  \multicolumn{1}{c|}{\textbf{0.953}} &
   \textbf{0.955}
   &
  \multicolumn{1}{c|}{\textbf{0.02}} &
   \textbf{8.90} \\ \hline
\end{tabular}%
}
\label{fig:experiment_1}
\end{figure}

Given the fundamental pitfalls described in Section~\ref{Section.2 Pitfalls} we focus on the challenges vital to clinical deployment --- conducting evaluations in the native image spaces. We pick four axes of algorithmic complexity: \textbf{(i)} voxel count (i.e. volume size), \textbf{(ii)} image spacing/anisotropy, \textbf{(iii)} target geometry (spherical vs irregular), and \textbf{(iv)} target size variation. We chose binary semantic segmentation tasks from the Medical Segmentation Decathlon (MSD) \cite{Antonelli2022TheDecathlon}, restricted by algorithmic fingerprints (Fig.~\ref{fig:algo_fingerprint}), both because they capture the axes of complexity described above and for reproducibility. For multi-sequence datasets, the sequence best visualising the target was used. Tasks include whole hippocampus (small volume), brain tumour core (T2w; medium volume, irregular shape, isotropic), whole pancreas (large volume and target), whole prostate (T2w; spherical target, highly anisotropic), and lung lesion (largest component; small target in large volume). All simulations use one point per iteration per class (as it is a SAM-Med3D constraint), uniform randomly sampled from false-negative foreground and background regions computed between predictions and reference annotations. Simulations are fully interactive with 100 editing steps; SegVol uses atomic inference (Section \ref{Section3:Methods}). Per-sample evaluation uses Dice and NSD (MSD tolerances) per iteration, and interaction count normalised AUCs (nAUC). These metrics are reported with dataset-wide medians. nnU-Net \cite{Isensee2021NnU-Net:Segmentation} models were also trained with 2-fold cross-validation (due to limited compute) with optimal configuration selection. We then used the segmentation performance of nnU-Net trained on the full dataset as a target performance level. We chose task-specific lower quartile Dice score rather than mean performance due to nnU-Net's asymmetric, heavy-tailed Dice distributions. With this target performance, we then estimate the number of interactions (NoI) for convergence on a per-sample Dice basis; the median NoI is then normalised by interaction count (nNoI). The fraction of samples that did not reach the performance target (NoF) is also reported. \\
\noindent\textbf{Experiment 1 - Image voxel count:} As shown in Fig.~\ref{fig:experiment_1}, 2D methods trail 3D ones in low-interaction settings across tasks. However, for small-volume hippocampus, both 2D models exceed the best 3D peak (SegVol) within 40 interactions. SAM2 achieves the highest final Dice, NSD, and nAUC scores, and is the only method to ever converge. For medium volumes (tumour core) we observed that 2D methods need more interactions to surpass the best 3D method (SegVol). SAM-Med2D obtains best peak and final Dice and NSD scores but SegVol performs better for the first 60 interactions with superior nAUC, nNoI, and NoF scores highlighting its rapid convergence. In large volumes (pancreas), SegVol performs best across iterations, maintaining strong leads in all metrics; converging rapidly and consistently with its zoom-out zoom-in mechanism (strongest nNoI and NoF scores). SAM-Med3D performs poorly due to lossy point mapping, retaining only foreground points and randomly sampling from point-centered resampled boxes for inference. Also, we note that SegVol improves initially but degrades with further interactions across tasks. This is pronounced in smaller volumes, suggesting densely packed points may induce undersegmentation. \\

\begin{figure}[t!]
\caption{\textbf{Top:} Median Dice and NSD across refinement simulations for the whole prostate and brain tumour core tasks. Axes of complexity: Highly anisotropic versus isotropic images, and spherical targets versus irregularly shaped targets. \textbf{Bottom:} Summary metrics for experiments 2 \& 3; all metrics but NoF report dataset medians, NoF reports percentages. Bold indicates best metric per task.}
\centering
\includegraphics[width=\linewidth]{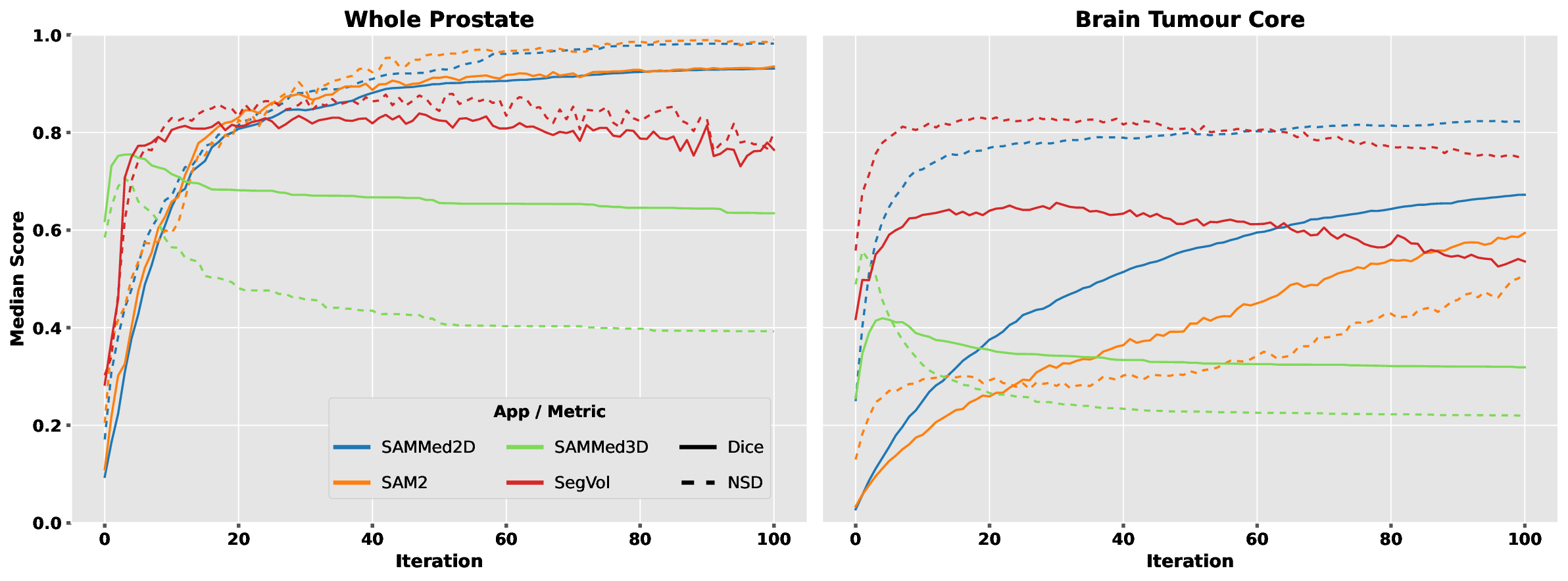}
\centering
\resizebox{0.8\textwidth}{!}{%
\begin{tabular}{|c|c|ccc|ccc|cc|}
\hline
\multirow{2}{*}{Task} &
  \multirow{2}{*}{Algorithm} &
  \multicolumn{3}{c|}{Dice} &
  \multicolumn{3}{c|}{NSD} &
  \multicolumn{2}{c|}{NoI} \\ \cline{3-10} 
 &
   &
  \multicolumn{1}{c|}{Init} &
  \multicolumn{1}{c|}{Iter. 100} &
  nAUC &
  \multicolumn{1}{c|}{Init} &
  \multicolumn{1}{c|}{Iter. 100} &
  nAUC &
  \multicolumn{1}{c|}{nNoI} &
  NoF \\ \hline
\multirow{4}{*}{Prostate} &
  SAM2 &
  \multicolumn{1}{c|}{0.110} &
  \multicolumn{1}{c|}{\textbf{0.935}} &
  \textbf{0.843} &
  \multicolumn{1}{c|}{0.205} &
  \multicolumn{1}{c|}{\textbf{0.991}} &
  \textbf{0.875} &
  \multicolumn{1}{c|}{\textbf{0.277}} &
  \textbf{0.0} \\ \cline{2-10} 
 &
  SAM-Med2D &
  \multicolumn{1}{c|}{0.094} &
  \multicolumn{1}{c|}{0.931} &
   0.835 &
  \multicolumn{1}{c|}{0.171} &
  \multicolumn{1}{c|}{0.982} &
   0.873
   &
  \multicolumn{1}{c|}{0.391} &
   3.13
   \\ \cline{2-10} 
 &
  SAM-Med3D &
  \multicolumn{1}{c|}{\textbf{0.619}} &
  \multicolumn{1}{c|}{0.634} &
   0.665 &
  \multicolumn{1}{c|}{\textbf{0.585}} &
  \multicolumn{1}{c|}{0.393} &
   0.443
   &
  \multicolumn{1}{c|}{1.0} &
   84.4
   \\ \cline{2-10} 
 &
  SegVol &
  \multicolumn{1}{c|}{0.283} &
  \multicolumn{1}{c|}{0.765} &
   0.782
   &
  \multicolumn{1}{c|}{0.303} &
  \multicolumn{1}{c|}{0.799} &
   0.801
   &
  \multicolumn{1}{c|}{1.0} &
   59.4
   \\ \hline
\multirow{4}{*}{Brain Tumour} &
  SAM2 &
  \multicolumn{1}{c|}{0.034} &
  \multicolumn{1}{c|}{0.594} &
   0.396 &
  \multicolumn{1}{c|}{0.130} &
  \multicolumn{1}{c|}{0.509} &
   0.366
   &
  \multicolumn{1}{c|}{1.0} &
   55.8
   \\ \cline{2-10} 
 &
  SAM-Med2D &
  \multicolumn{1}{c|}{0.028} &
  \multicolumn{1}{c|}{\textbf{0.672}} &
   0.504
   &
  \multicolumn{1}{c|}{0.250} &
  \multicolumn{1}{c|}{\textbf{0.822}} &
   0.776
   &
  \multicolumn{1}{c|}{0.851} &
   43.8
   \\ \cline{2-10} 
 &
  SAM-Med3D &
  \multicolumn{1}{c|}{0.256} &
  \multicolumn{1}{c|}{0.319} &
   0.341
   &
  \multicolumn{1}{c|}{0.489} &
  \multicolumn{1}{c|}{0.219} &
  0.256
   &
  \multicolumn{1}{c|}{1.0} &
   77.9 \\ \cline{2-10} 
 &
  SegVol &
  \multicolumn{1}{c|}{\textbf{0.418}} &
  \multicolumn{1}{c|}{0.536} &
   \textbf{0.595}
   &
  \multicolumn{1}{c|}{\textbf{0.558}} &
  \multicolumn{1}{c|}{0.750} &
   \textbf{0.791}
   &
  \multicolumn{1}{c|}{\textbf{0.060}} &
   \textbf{33.7}
   \\ \hline
\end{tabular}%
}
\label{fig:experiment_2_3}
\end{figure}

\noindent\textbf{Experiment 2 - Image anisotropy:} Fig.~\ref{fig:experiment_2_3} shows that 3D algorithms perform better at initialisation for both tasks. But, on slab-like images (prostate), 2D methods improve rapidly, becoming competitive after a few interactions. Around 30 iterations, 2D methods surpass the best 3D method (SegVol) in Dice and NSD, with fewer convergence failures (NoF). SAM2 achieves the best overall performance except in initialisation metrics, with no convergence failures. For isotropic images (tumour core), 2D methods are less competitive. SAM-Med2D achieves highest final and peak Dice and NSD but only surpasses SegVol after  60 edits which consistently converges rapidly across most metrics. \\ 

\noindent\textbf{Experiment 3 - Target geometry:} Fig.~\ref{fig:experiment_2_3} shows that SAM-Med3D has its best performance across any task for whole prostate. Considerable degradation on tumour core is due to SAM-Med3D's lossy prompt mapping preventing correction of oversegmentations; vital for non-coarse, non-spherical targets. SAM2 delivers best overall performance on whole prostate but performs worse than SAM-Med2D on tumour core, indicating medical-domain adaptation for irregular, ambiguous targets is important for points. SegVol's strong peak Dice and NSD metrics; and best nAUC, nNoI and NoF metrics on tumour core indicate volumetric context is key to rapid, consistent convergence on irregular structures. \\

\begin{figure}[t!]
\caption{\textbf{Top:} Median Dice and NSD across refinement simulations for variation in target size relative to the image volume (lung lesion versus whole pancreas). \textbf{Bottom:} Summary metrics for experiment 4; all metrics but NoF report dataset medians, NoF reports percentages. Bold indicates best metric per task.}
\includegraphics[width=\linewidth]{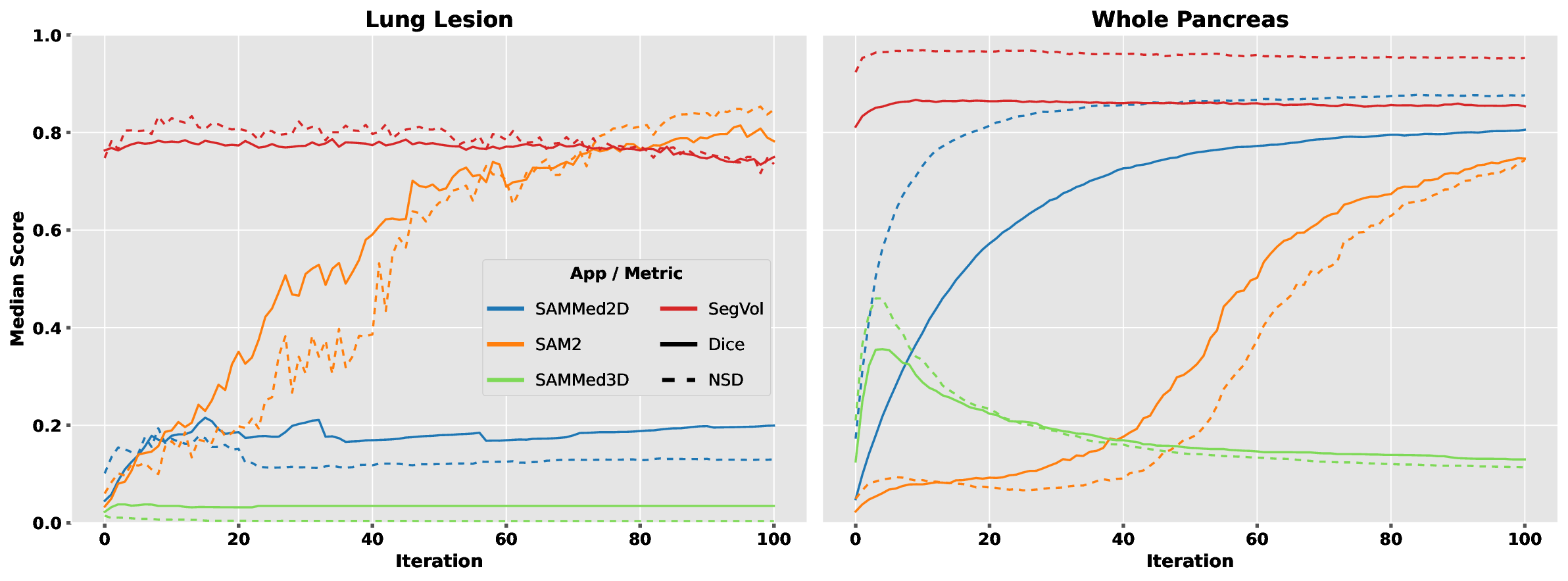}
\centering
\resizebox{0.8\textwidth}{!}{%
\begin{tabular}{|c|c|ccc|ccc|cc|}
\hline
\multirow{2}{*}{Task} &
  \multirow{2}{*}{Algorithm} &
  \multicolumn{3}{c|}{Dice} &
  \multicolumn{3}{c|}{NSD} &
  \multicolumn{2}{c|}{NoI} \\ \cline{3-10} 
 &
   &
  \multicolumn{1}{c|}{Init} &
  \multicolumn{1}{c|}{Iter. 100} &
  nAUC &
  \multicolumn{1}{c|}{Init} &
  \multicolumn{1}{c|}{Iter. 100} &
  nAUC &
  \multicolumn{1}{c|}{nNoI} &
  NoF \\ \hline
\multirow{4}{*}{Lung Lesion} &
  SAM2 &
  \multicolumn{1}{c|}{0.033} &
  \multicolumn{1}{c|}{\textbf{0.782}} &
  0.572 &
  \multicolumn{1}{c|}{0.060} &
  \multicolumn{1}{c|}{\textbf{0.847}} &
  0.543 &
  \multicolumn{1}{c|}{0.386} &
  \textbf{15.9} \\ \cline{2-10} 
 &
  SAM-Med2D &
  \multicolumn{1}{c|}{0.045} &
  \multicolumn{1}{c|}{0.199} &
   0.211 &
  \multicolumn{1}{c|}{0.102} &
  \multicolumn{1}{c|}{0.130} &
   0.141
   &
  \multicolumn{1}{c|}{1.0} &
   76.2 \\ \cline{2-10} 
 &
  SAM-Med3D &
  \multicolumn{1}{c|}{0.023} &
  \multicolumn{1}{c|}{0.035} &
   0.034 &
  \multicolumn{1}{c|}{0.015} &
  \multicolumn{1}{c|}{0.004} &
   0.007
   &
  \multicolumn{1}{c|}{1.0} &
   96.8 \\ \cline{2-10} 
 &
  SegVol &
  \multicolumn{1}{c|}{\textbf{0.763}} &
  \multicolumn{1}{c|}{0.750} &
   \textbf{0.768}
   &
  \multicolumn{1}{c|}{\textbf{0.748}} &
  \multicolumn{1}{c|}{0.736} &
   \textbf{0.777} 
   &
  \multicolumn{1}{c|}{\textbf{0.010}} &
   \textbf{15.9}
   \\ \hline
\multirow{4}{*}{Pancreas} &
  SAM2 &
  \multicolumn{1}{c|}{0.023} &
  \multicolumn{1}{c|}{0.746} &
   0.375
   &
  \multicolumn{1}{c|}{0.050} &
  \multicolumn{1}{c|}{0.743} &
   0.321
   &
  \multicolumn{1}{c|}{1.0} &
   75.1 \\ \cline{2-10} 
 &
  SAM-Med2D &
  \multicolumn{1}{c|}{0.049} &
  \multicolumn{1}{c|}{0.806} &
   0.674
   &
  \multicolumn{1}{c|}{0.173} &
  \multicolumn{1}{c|}{0.876} &
   0.824
   &
  \multicolumn{1}{c|}{1.0} &
   63.3 \\ \cline{2-10} 
 &
  SAM-Med3D &
  \multicolumn{1}{c|}{0.126} &
  \multicolumn{1}{c|}{0.130} &
   0.189
   &
  \multicolumn{1}{c|}{0.210} &
  \multicolumn{1}{c|}{0.114} &
   0.184
   &
  \multicolumn{1}{c|}{1.0} &
   98.9 \\ \cline{2-10} 
 &
  SegVol &
  \multicolumn{1}{c|}{\textbf{0.812}} &
  \multicolumn{1}{c|}{\textbf{0.854}} &
   \textbf{0.857}
   &
  \multicolumn{1}{c|}{\textbf{0.924}} &
  \multicolumn{1}{c|}{\textbf{0.953}} &
   \textbf{0.955}
   &
  \multicolumn{1}{c|}{\textbf{0.020}} &
   \textbf{8.90} \\ \hline
\end{tabular}%
}
\label{fig:experiment_4}
\end{figure}

\noindent\textbf{Experiment 4 - Target size: } As shown in Fig.~\ref{fig:experiment_4}, for lung lesion (small target), SAM-Med2D performs poorly throughout. This is likely due to removing small targets from training data, improving in larger pancreas targets. SAM-Med3D struggles on both tasks, especially small targets, due to noisy point mapping to model-space. For lung lesion, although SAM2 reaches best peak Dice and NSD, SegVol converges faster with superior nAUC and nNoI and comparable failure rates. For the larger pancreas targets, SAM2 converges slower, while SegVol still achieves best performance across all metrics. The sizeable margin from the second best method on pancreas (SAM-Med2D) for Dice, NSD and NoF indicates volumetric context is vital for performance. Lastly, SegVol's robustness across target-size variation outlines the utility of dynamically adapting to target size.

\section{Conclusions and Future Work}

The proposed framework and evaluation highlights several takeaways: \textbf{(i)} minimising information loss when processing prompts and using adaptive zooming strategies enables robustness across variation in volume and target sizes \textbf{(ii)} adaptive zooming strategies enable rapid convergence \textbf{(iii)} performance can degrade if prompting behaviour or budgets diverge from that in training; \textbf{(iv)} 2D methods can excel on image-slabs with coarse targets, but volumetric context aids in large or irregularly shaped targets, even for out-of-distribution modalities (e.g., SegVol for tumour core); \textbf{(v)} SAM2 can perform well without medical-domain training but suffers on tissue-ambiguous targets like brain tumours when using points. Future work will explore rank stability of metrics, expand the pool of integrated algorithms and broaden evaluation tasks. Packaging algorithms as applications also enables future user studies to design user-realistic prompting simulations, akin to Antonov et al. \cite{Antonov2024RClicks:Segmentation}. Future user-studies assessing prompt placement time across different prompt types and target geometries will also be vital for effort-estimation. Lastly, future work will need to ensure that validation datasets were never used for model pre-training; this is a limitation of the proposed work and most works that build on foundation models.

\begin{credits}
\subsubsection{\ackname} Parhom Esmaeili is supported by the Engineering and Physical Sciences Research Council iCASE (grant number EP/Y528572/1) and co-funded by Siemens Healthineers. Virginia Fernandez is supported by NHS England.

\subsubsection{\discintname}
The authors declare no competing interests to report.
\end{credits}

%
%
%
%




\bibliography{Paper-0013}

\begin{thebibliography}{10}
\providecommand{\url}[1]{\texttt{#1}}
\providecommand{\urlprefix}{URL }
\providecommand{\doi}[1]{https://doi.org/#1}

\bibitem{Antonelli2022TheDecathlon}
Antonelli, M., Reinke, A., Bakas, S., et~al.: {The Medical Segmentation Decathlon}. Nature Communications  \textbf{13}(1), ~4128 (7 2022). \doi{10.1038/s41467-022-30695-9}

\bibitem{Antonov2024RClicks:Segmentation}
Antonov, A., Moskalenko, A., Shepelev, D., et~al.: {RClicks: Realistic Click Simulation for Benchmarking Interactive Segmentation}. In: Globerson, A., Mackey, L., Belgrave, D., Fan, A., Paquet, U., Tomczak, J., Zhang, C. (eds.) Advances in Neural Information Processing Systems. pp. 127673--127710. Curran Associates Inc, Vancouver (2024)

\bibitem{Boykov2001InteractiveImages}
Boykov, Y., Jolly, M.P.: {Interactive graph cuts for optimal boundary {\&} region segmentation of objects in N-D images}. In: Proceedings Eighth IEEE International Conference on Computer Vision. ICCV 2001. pp. 105--112. IEEE Comput. Soc, Vancouver (2001). \doi{10.1109/ICCV.2001.937505}

\bibitem{Cheng2023SAM-Med2D}
Cheng, J., Ye, J., Deng, Z., Chen, J., Li, T., Wang, H., Su, Y., Huang, Z., Chen, J., Jiang, L., Sun, H., He, J., Zhang, S., Zhu, M., Qiao, Y.: {SAM-Med2D} (8 2023)

\bibitem{Du2024SegVol:Segmentation}
Du, Y., Bai, F., Huang, T., Zhao, B.: {SegVol: Universal and Interactive Volumetric Medical Image Segmentation}. In: Globerson, A., Mackey, L., Belgrave, D., Fan, A., Paquet, U., Tomczak, J., Zhang, C. (eds.) Advances in Neural Information Processing Systems. pp. 110746--110783. Curran Associates Inc, Vancouver (2024)

\bibitem{Fedorov20123DNetwork}
Fedorov, A., Beichel, R., Kalpathy-Cramer, J., et~al.: {3D Slicer as an image computing platform for the Quantitative Imaging Network}. Magnetic Resonance Imaging  \textbf{30}(9),  1323--1341 (11 2012). \doi{10.1016/j.mri.2012.05.001}

\bibitem{Gatidis2022ALesions}
Gatidis, S., Hepp, T., Fr{\"{u}}h, M., et~al.: {A whole-body FDG-PET/CT Dataset with manually annotated Tumor Lesions}. Scientific Data  \textbf{9}(1), ~601 (10 2022). \doi{10.1038/s41597-022-01718-3}

\bibitem{Isensee2021NnU-Net:Segmentation}
Isensee, F., Jaeger, P.F., Kohl, S.A.A., Petersen, J., Maier-Hein, K.H.: {nnU-Net: a self-configuring method for deep learning-based biomedical image segmentation}. Nature Methods  \textbf{18}(2),  203--211 (2 2021). \doi{10.1038/s41592-020-01008-z}

\bibitem{Isensee2025NnInteractive:Segmentation}
Isensee, F., Rokuss, M., Kr{\"{a}}mer, L., et~al.: {nnInteractive: Redefining 3D Promptable Segmentation}. arXiv preprint  (3 2025)

\bibitem{Kass1988Snakes:Models}
Kass, M., Witkin, A., Terzopoulos, D.: {Snakes: Active contour models}. International Journal of Computer Vision  \textbf{1}(4),  321--331 (1 1988). \doi{10.1007/BF00133570}

\bibitem{Kirillov2023SegmentAnything}
Kirillov, A., Mintun, E., Ravi, N., Mao, H., Rolland, C., Gustafson, L., Xiao, T., Whitehead, S., Berg, A.C., Lo, W.Y., Doll{\'{a}}r, P., Girshick, R.: {Segment Anything}. arXiv preprint  (4 2023)

\bibitem{deKok2023ARegulation}
de~Kok, J.W.T.M., de~la Hoz, M.A.A., de~Jong, Y., et~al.: {A guide to sharing open healthcare data under the General Data Protection Regulation}. Scientific Data  \textbf{10}(1), ~404 (6 2023). \doi{10.1038/s41597-023-02256-2}

\bibitem{Lenchik2019AutomatedReview}
Lenchik, L., Heacock, L., Weaver, A.A., et~al.: {Automated Segmentation of Tissues Using CT and MRI: A Systematic Review}. Academic Radiology  \textbf{26}(12),  1695--1706 (12 2019). \doi{10.1016/j.acra.2019.07.006}

\bibitem{Li2024Prism:Prompts}
Li, H., Liu, H., Hu, D., Wang, J., Oguz, I.: {Prism: A promptable and robust interactive segmentation model with visual prompts}. In: Linguraru, M.G., Dou, Q., Feragen, A., Giannarou, S., Glocker, B., Lekadir, K., Schnabel, J.A. (eds.) International Conference on Medical Image Computing and Computer-Assisted Intervention - MICCAI 2024. pp. 389--399. Springer, Marrakesh (10 2024)

\bibitem{Maier-Hein2024MetricsValidation}
Maier-Hein, L., Reinke, A., Godau, P., et~al.: {Metrics reloaded: recommendations for image analysis validation}. Nature Methods  \textbf{21}(2),  195--212 (2 2024). \doi{10.1038/s41592-023-02151-z}

\bibitem{Maleki2025AnalysisMRI}
Maleki, N., Amiruddin, R., Moawad, A.W., et~al.: {Analysis of the MICCAI Brain Tumor Segmentation -- Metastases (BraTS-METS) 2025 Lighthouse Challenge: Brain Metastasis Segmentation on Pre- and Post-treatment MRI}. arXiv preprint  (5 2025)

\bibitem{Ravi2024SAMVideos}
Ravi, N., Gabeur, V., Hu, Y.T., et~al.: {SAM 2: Segment Anything in Images and Videos}. arXiv preprint  (8 2024)

\bibitem{Ulrich2025RadioActive:Benchmark}
Ulrich, C., Wald, T., Tempus, E., Rokuss, M., Jaeger, P.F., Maier-Hein, K.: {RadioActive: 3D Radiological Interactive Segmentation Benchmark}. arXiv preprint  (3 2025)

\bibitem{Wang2023SAM-Med3D}
Wang, H., Guo, S., Ye, J., Deng, Z., Cheng, J., Li, T., Chen, J., Su, Y., Huang, Z., Shen, Y., Fu, B., Zhang, S., He, J., Qiao, Y.: {SAM-Med3D}. arXiv preprint  (10 2023)

\bibitem{Yushkevich2006User-guidedReliability}
Yushkevich, P.A., Piven, J., Hazlett, H.C., Smith, R.G., Ho, S., Gee, J.C., Gerig, G.: {User-guided 3D active contour segmentation of anatomical structures: Significantly improved efficiency and reliability}. NeuroImage  \textbf{31}(3),  1116--1128 (7 2006). \doi{10.1016/j.neuroimage.2006.01.015}

\end{thebibliography}
\bibliographystyle{splncs04}
\end{document}